\RequirePackage{fix-cm}
\documentclass[smallextended]{svjour3}       
\smartqed  

\usepackage{url}
\usepackage{hyperref}
\usepackage{lipsum}
\usepackage[table]{xcolor}
\usepackage{calrsfs} 
\DeclareMathAlphabet{\pazocal}{OMS}{zplm}{m}{n}
\usepackage[ruled, linesnumbered]{algorithm2e} 
\usepackage{graphicx}
\usepackage[utf8]{inputenc}
\usepackage{graphics, subfig} 
\usepackage{amsmath} 
\usepackage{amssymb}  
\usepackage{mathtools}
\DeclarePairedDelimiterX{\norm}[1]{\lVert}{\rVert}{#1}
\SetCommentSty{mycommfont}
\SetEndCharOfAlgoLine{}

\makeatletter
\newcommand{\algorithmfootnote}[2][\scriptsize]{%
  \let\old@algocf@finish\@algocf@finish
  \def\@algocf@finish{\old@algocf@finish
    \leavevmode\rlap{\begin{minipage}{\linewidth}
    #1#2
    \end{minipage}}%
  }%
}
\makeatother

\journalname{Multimedia Tools and Applications}

\newcommand{\ourMethod}{Concept-Drift-Aware Federated Averaging}
\newcommand{\ourAcronym}{CDA-FedAvg}
\begin{document}

\title{Concept drift detection and adaptation \\ for federated and continual learning}

\author{Fernando E. Casado \and 
        Dylan Lema \and
        Marcos F. Criado \and
        Roberto Iglesias \and 
        Carlos V. Regueiro \and 
        Sen\'en Barro
}

\institute{Fernando E. Casado, Dylan Lema, Marcos F. Criado, Roberto Iglesias,  and Sen\'en Barro \at 
        CiTIUS (Centro Singular de Investigaci\'on en Tecnolox\'ias Intelixentes), Universidade de Santiago de Compostela, 15782 Santiago de Compostela, Spain \\
        \email{\{fernando.estevez.casado, dylan.lema, marcos.criado, roberto.iglesias.rodriguez,  senen.barro\}@usc.es}
        \and
        Carlos V. Regueiro \at
        CITIC, Computer Architecture Group, \\ Universidade da Coru\~na, 15071 A Coru\~na, Spain \\
        \email{carlos.vazquez.regueiro@udc.es}}
        
\date{ }

\maketitle              

\begin{abstract}
Smart devices, such as smartphones, wearables, robots, and others, can collect vast amounts of data from their environment. This data is suitable for training machine learning models, which can significantly improve their behavior, and therefore, the user experience. Federated learning is a young and popular framework that allows multiple distributed devices to train deep learning models collaboratively while preserving data privacy. Nevertheless, this approach may not be optimal for scenarios where data distribution is non-identical among the participants or changes over time, causing what is known as \textit{concept drift}. Little research has yet been done in this field, but this kind of situation is quite frequent in real life and poses new challenges to both continual and federated learning. Therefore, in this work, we present a new method, called \ourMethod{}~(\ourAcronym{}). Our proposal is an extension of the most popular federated algorithm, Federated Averaging~(FedAvg), enhancing it for continual adaptation under concept drift. We empirically demonstrate the weaknesses of regular FedAvg and prove that \ourAcronym{} outperforms it in this type of scenario.

\keywords{Federated learning \and Continual learning \and Nonstationarity \and Concept drift \and Federated Averaging \and Catastrophic forgetting \and Rehearsal}
\end{abstract}
%

\section{Introduction}
\label{sec:intro}

Over the last few decades, our society has experienced a technological explosion which, among other things, has gradually surrounded us with smart devices that are part of our daily lives. We are talking about smartphones, of course, but also wearables, ``things'' from the Internet of Things (IoT), service robots, etc. In short, multifunctional devices with cutting-edge technology that allows a great number of applications from all human domains: health, sport, education, banking, etc.
The growing amount of data that these devices can collect, together with a good intercommunication between them, enables the possibility of integrating machine learning models that evolve and adapt to improve their behaviour. 

A good way to learn from the data collected on the devices would be by consensus, in which a global model is built from the partial data provided by each participant. 
This learning process can be addressed either in a centralized manner, employing traditional and offline server-based architectures, or in a decentralized way, following approaches such as \emph{federated learning}~(FL)~\cite{konevcny2015federated,mcmahan2016federated,li2019federated}. Centralized methods involve collecting and uploading data from all the participants, also known as  \emph{clients} or \emph{agents}, to a cloud-based server or data center, where it is processed. Decentralized methods, instead, aim to solve local sub-problems on the devices in a distributed and parallel way, and then usually combine the local solutions in the cloud.

Centralized solutions are undoubtedly still the most common option nowadays. 
Nevertheless, applying cloud-centric approaches to learn from the data collected on the devices involves important issues:
\begin{itemize}
    \item \textbf{Scalability}. This concerns both storage and communication costs, but also computing speeds. By having a central node acting as a server, there is always a risk that this will be a bottleneck. The more responsibilities the server has, the higher the risk.
    Transferring large amounts of data over the network take a long time, and communication may be a continuous overhead. Similarly, central processing can take much more time than parallel computing in the clients.

    \item \textbf{Data privacy and sensitivity}. Central data collection also puts user privacy at risk. The information collected on the devices may be sensitive. Therefore, over the last few years, several governments around the world have implemented new legislation to protect the data privacy, limiting data sending and storage only to what is consented by the consumer and absolutely necessary for processing. Examples of this are the European Commission's General Data Protection Regulation (GDPR)~\cite{custers2019eu} or the Consumer Privacy Bill of Rights in the US~\cite{gaff2014privacy}. 
    
    \item \textbf{Adaptability}. In machine learning, it is common to assume that data is stationary and IID (independent and identically distributed). However, in most real-world situations, the underlying distribution of data changes between the different participants and also evolves over time. Heterogeneity among participants cannot be solved by learning a single centralized model. On the other hand, changes in data over time have been extensively studied in the literature and are commonly referred to as \emph{concept drift}~\cite{lu2018,lesort2020continual}. If a concept drift happens, the patterns learned so far may not be relevant anymore, leading to poor model performance. Therefore, in this kind of situation, it is desirable to detect when these drifts occur in order to adapt to them. Although several solutions have been proposed for handling concept drift, none of them has been specifically designed for a multi-device setting. In this scenario, centralized management of concept drift could increase bandwidth, storage, and computational costs in the server, which brings us back to the first issue, scalability.
\end{itemize}

Due to these problems, decentralized solutions are best suited for learning from distributed devices. In this way, \emph{federated learning} (FL)~\cite{konevcny2015federated,mcmahan2016federated,li2019federated} has been positioned in recent years as the reference for distributed and collaborative machine learning, dealing with user privacy and scalability. 
Nevertheless, literature on FL has paid little attention to adaptability, particularly in the temporal dimension.
In this regard, \emph{continual learning} (CL)~\cite{lesort2020continual,parisi2019}, is the state-of-the-art paradigm for addressing this problem of data nonstationarity and adaptability over time. In order to achieve adaptive models, CL deals with two conflicting objectives: retaining previously learned knowledge that is still relevant, and replacing any obsolete knowledge with current information. This is usually known as the \emph{stability-plasticity dilemma}~\cite{grossberg1988}. 

In this work, we introduce an extension of the most widely used FL algorithm, \emph{Federated Averaging} (FedAvg)~\cite{mcmahan2016federated}, adapting it for learning under concept drift. In particular, we extend the algorithm to deal with \emph{continual single-task problems}. That is, we focus on the scenario where all the client devices share the same goal, but the underlying joint distribution of data might be non-IID (independent and identically distributed) among them, and can also change in unforeseen ways over time (concept drift). We call our new method \emph{\ourMethod{}}~(\ourAcronym{}). In this article we also describe the performance of our method when it has been tested for the task of human activity recognition (HAR) using smartphones. In this case, we have used a dataset created from the data collected by the mobile devices of 10 different users. 
Our experimental results show the weaknesses of regular FedAvg in this kind of continual setting and a remarkable improvement in the learning ability of \ourAcronym{} over FedAvg. We also prove a reduction in storage, communication, and computational costs in the devices. This paper is an extension of the work originally presented in the 21\textsuperscript{st} International Workshop of Physical Agents (WAF 2020)~\cite{casado2020concept}.

The rest of this paper is organized as follows: Section~\ref{sec:related-work} reviews the state of the art on continual and multi-device learning. Section~\ref{sec:problem-definition} provides a formal definition of concept drift and outlines the fundamentals of federated learning. In Section~\ref{sec:drift}, \ourAcronym{} is introduced. Section~\ref{sec:exp-results} presents the experimental results. Finally, Section~\ref{sec:conclusions} draws our main conclusions and future work.

\section{Related Work}
\label{sec:related-work}

Continual learning~\cite{lesort2020continual,parisi2019} is a machine learning paradigm focused on building adaptive models over time. This approach seeks to smoothly update the predictor to take into account different data distributions and tasks but still being able to re-use and retain useful knowledge during the time. It is highly inspired by the human learning process, as people learn to perform numerous tasks over their lifespan, making use of past knowledge to learn about new concepts without forgetting the previous ones. CL deals with high and realistic time scales where data becomes available only during the time, and past information is not always accessible.

Concept drift is one of the two main challenges ---together with \emph{catastrophic forgetting}~\cite{french1999catastrophic}---  currently being addressed by CL literature~\cite{lesort2020continual}. When the distribution is not stationary, a shift into the data stream is observed. If no external information regarding this shift is available, we must employ some other strategy to detect it, and also fix it. An undetected shift in the data distribution will lead to a downgrade in the model performance. These unpredictable changes in the data distribution over time are what is commonly known as \emph{concept drift}.
Handling concept drift also involves managing memories of past concepts, which can be saved in different manners: as raw data, as representations, as model weights, etc. An efficient memory management strategy should only save important information, as well as be able to transfer knowledge and skills to future tasks.

Federated learning~\cite{konevcny2015federated,mcmahan2016federated,li2019federated} is a distributed machine learning framework under differential privacy. Basically, it consists of a local learning stage in the devices, and a global parameter aggregation in a central server in the cloud. Usually, there is a shared model, which is a Deep Neural Network (DNN). The learnable parameters of this model are initialized on the server. In each federated round, the clients receive the current parameter set from the server, perform stochastic gradient descent (SGD) on their local datasets, and send back the gradients. Guarantees about the lack of sensitive information of the gradient have been widely studied to preserve client privacy~\cite{kairouz2019advances}. After that, the contributions of each device are aggregated on the server, thus updating the model. The most popular FL algorithm is \emph{Federated Averaging} (FedAvg)~\cite{mcmahan2016federated}. The different challenges this framework faces, such as the communication costs and data statistical heterogeneity, have also been given significant attention in the literature  \cite{li2020federated}. FL has been widely applied to solve complex classification and regression tasks, such as object recognition and movement detection in images, predictive text on the smartphone keyboard, or patient mortality prediction and hospital stay time~\cite{li2019federated,aledhari2020federated}.

As we exposed in Section~\ref{sec:intro}, three main challenges must be addressed for \emph{multi-device learning}: (1)~scalability, (2)~data privacy, and (3)~adaptability. 
Federated learning seems, at present, the most suitable approach for this context, since it successfully tackles the first 2 challenges. However, continual learning is undoubtedly the paradigm by reference as far as adaptability is concerned. 
Addressing both FL and CL in an orthogonal way can be the best way to tackle at the same time all the three challenges. Nevertheless, little work has yet been done in this regard.

Some authors have evaluated Federated Averaging on non-IID scenarios, analyzing the impact of non-identical client data partitions. It is the case of McMahan et al.~\cite{mcmahan2016federated}, that synthesize pathological non-identical user splits from the MNIST dataset, or Zhao et al.~\cite{zhao2018federated}, that does the equivalent using the CIFAR-10 dataset. They demonstrate that FedAvg on non-identical clients still converges to high accuracies, though taking more rounds than identical clients. Other authors study more realistic client data distributions. For instance, Caldas et al.~\cite{caldas2018leaf} use the Extended MNIST dataset with partitions over writers of the digits, rather than simply partitioning over digit class. 
Finally, there are some authors that, instead of assuming that the global model should be identical for all clients, have proposed a local personalization for each participant. In this way, Li et al. \cite{li2018federated} allow certain variation between the model of each client and the model obtained with FedAvg. On the other hand, Deng et al. \cite{deng2020adaptive} keep track of each clients' local model and take in into account for adaptation.

All the aforementioned works focus on analyzing the impact of non-identical client data partitions in FedAvg performance. However, little effort has been made to evaluate the effect of nonstationary, time-shifting data streams. Yoon et al.~\cite{yoon2020federated} propose a method called \emph{Federated Continual Learning with Weighted Inter-client Transfer}, FedWeIt, which additively decomposes the weights of the network into two separated sets: globally shared parameters and sparse task-specific parameters. Hence, they are capable of performing inter-client knowledge transfer and prevent inter-client interference. They validate their approach on several settings, including continual multi-incremental-task scenarios. When comparing their method with existing federated baselines, they show significantly higher accuracy and reduced communication costs. However, this work still does not explicitly detect changes in data distribution.

%

In the present paper, the original FedAvg algorithm is redesigned in order to face concept drift and continual learning.
As will be discussed below, there are significant benefits from drift detection. In short, we can say that it helps us answering two key questions~\cite{casado2020collaborative}: \emph{what} to learn and \emph{when} to learn it. In this article, we tackle the implications of these unpredictable changes, focusing on a single-task scenario in which different changes occur over time.

\section{Concept Drift in Federated Settings: Problem~Definition}
\label{sec:problem-definition}

In standard federated settings, the learning process involves multiple rounds of local learning and global aggregation. At each round $r$, each client $j\in \{1, \dots,C\}$ and the server $s$ perform two different learning stages: \emph{local parameter update} and \emph{global parameter aggregation}. 

The learnable parameters of the model (weights and biases) are initialized on the server. At the beginning of each round, a random subset of clients, $\pazocal{C}_{r}~\subseteq~\{1,2, \dots, C\}$, of size $m = |\pazocal{C}_{r}|$, is selected. The server sends the current global algorithm state to each of these clients, i.e., the current model parameters.  In the local parameter update step, each client $j$ performs stochastic gradient descent (SGD) on its local dataset and sends the updated parameters $\mathbf{w}^j_{r}$ to the server. 
Then, the server aggregates the parameters $\mathbf{w}^j_{r}$ sent from all the clients in $\pazocal{C}^{r}$ into a single parameter. The way in which this aggregation is carried out is usually the average, although there are variations depending on the aggregation method used.
In the particular case of Federated Averaging (FedAvg)~\cite{mcmahan2016federated}, the local parameters from each client are aggregated applying a weighted average:
\begin{equation*}
    \mathbf{w}^G_{r} \leftarrow \sum_{j=1}^{m} \frac{n^j}{N} \mathbf{w}^j_{r},
\end{equation*}
where $N$ is the total number of data instances at each round, and $n^j$ is the number of instances from client $j$. Then, the updated model is sent back to the devices, a new random subset of clients is selected, and a new training~round~starts. Algorithms~\ref{alg:fedavg}~and~\ref{alg:fedavg-client} show the pseudocode of FedAvg.

\SetKwInOut{kwInput}{Input} 
\SetKwInOut{kwOutput}{Output}
\begin{algorithm}[!htb] 
            \kwInput{Number of learning rounds $R$,  number of participants $m$ per round, local minibatch size $B$, number of local epochs $E$,  and learning rate $\eta$.} 
            \kwOutput{Global model $\mathbf{w}^G$.} 
		    Initialize $\mathbf{w}_0^G$ \\
			\For{$r \leftarrow 1$ \KwTo $R$}{
			    Randomly choose a subset $\pazocal{C}_{r}$ of $m$ participants from $\{1,2, \dots, C\}$ \\
				\For{\textnormal{each participant} $j \in \pazocal{C}_{r}$ \textnormal{parallely}}{
				   $\mathbf{w}_{r}^j \leftarrow \texttt{localTraining[j]}(\mathbf{w}_{r-1}^G, B, E, \eta)$  \tcp*{Local update (Algorithm \ref{alg:fedavg-client})} 
				}
				$\mathbf{w}_{r}^G \leftarrow \sum_{j=1}^{m} \frac{n^j}{N}\mathbf{w}^j_{r}$ \tcp*{ Averaging aggregation} 
			}
	\caption{Federated Averaging, server aggregation. \label{alg:fedavg}}
\end{algorithm}
\vspace{-0.8cm}
\SetKwInOut{kwInput}{Input} 
\SetKwInOut{kwOutput}{Output}
\begin{algorithm}[!htb] 
            \kwInput{Last global model $\mathbf{w}_{r-1}^G$, local minibatch size $B$, number of local epochs $E$, and learning rate $\eta$.} 
            \kwOutput{Local update $\mathbf{w}_{r}$.} 
            
		    Split the local dataset into a set $\pazocal{B}$ of batches of size $B$ \\
			\For{$e \leftarrow 1$ \KwTo $E$}{
			    \For{\textnormal{batch} $b \in \pazocal{B}$}{
			        $\mathbf{w}_{r} \leftarrow \mathbf{w}_{r-1}^G - \eta \nabla l (\mathbf{w}_{r-1}^G; b)$ \tcp*{$\nabla l$ is the gradient of $l$ on $b$}
				}
			}
	\caption{Federated Averaging, local update (\texttt{localTraining}).\label{alg:fedavg-client}}
\end{algorithm}

That is the original FedAvg procedure proposed by McMahan et al. in 2016~\cite{mcmahan2016federated}. Nevertheless, as we will show in Section~\ref{sec:exp-results}, naive training of FedAvg in real-world situations can lead to catastrophic forgetting, because local data is by nature nonstationary and non-IID. When data distribution is not stationary, a concept drift is observed in the data stream. In the absence of external information about this drift, the model will have to detect and adapt to it on its own to avoid decreasing its performance.

Concept drift is a continual learning challenge~\cite{lesort2020continual}. Formally, we can define it as follows: Given a time period $[0,t]$, a set of samples which we will denote as $S_{0,t} = \{(X_0,y_0) \dots, (X_t,y_t)\}$, where $(X_i, y_i)$ is one observation or data sample. $X_i$ is the feature vector, $y_i$ is the label, and $S_{0,t}$ has a certain joint probability distribution $P_{t}(X,y)$. A concept drift can be defined as a change in the joint probability at timestamp $t$, i.e., $\exists t: P_{t}(X,y) \neq P_{t+1}(X,y)$.

We can categorize concept drift according to different criteria. Notice that the joint probability can be factorized as follows: $P(X,y) = P(X) \cdot P(y \mid X)$. Thus, we can make a first categorization of concept drift based on which factor from the previous equation is altered. In this way, we distinguish between two types of change~\cite{webb2016characterizing,gepperth2016incremental}: (1)~virtual,  and (2)~real.
\emph{Virtual concept drift} just refers to shifts in the input distribution, $P(X)$, and can easily occur (e.g., due to imbalanced classes over time), whereas \emph{real concept drift} is caused by novelty on data, which has an effect on posterior class probabilities, $P(y \mid X)$.
Apart from this two cases, concept drift can also happen when the task changes. In this case, the change does not take place in the data distribution, but in the goal pursued. We focus on the \emph{single-incremental-task scenario}~\cite{lesort2020continual}, that is, the task remains unchanged all along. Besides, the drifts considered in this work are provoked by changes on input data ($P(X)$), that is, \emph{virtual concept drifts}. The reason for this decision is that we want our drift detection method (Section~\ref{sec:drift-detection}) to be completely unsupervised, i.e., to work without needing the pattern labels.

Furthermore, it is also interesting to classify concept drift considering how the new joint distribution is different from the previous one. In this case we can discern four types of concept drift: (i) sudden, (ii) recurring, (iii) gradual, and (iv) incremental. We say a concept drift is \textit{sudden} if there is a timestamp that separates the old concept from the new one. In particular, this process can happen repeatedly  and even go back to the original concept, and in that case we say it is a \textit{recurring} concept drift. On the contrary, if the data from the new concept arises intertwined with the old concept at first, we name it \textit{gradual} concept drift. \textit{Incremental} concept drift takes place when data shifts smoothly between the concepts, and therefore the drift cannot be detected in a single instant but within a window of consecutive timestamps. As we will see afterwards, the method we propose deals both with \textit{gradual} and \textit{sudden} drifts, and hence with \textit{recurring} ones too. The case of \textit{incremental} drift is more subtle, as our method could detect it in certain situations depending on some hyperparameters of the algorithm (see Section~\ref{sec:drift-detection}).

We now extend the conventional concept drift definition to the federated learning setting, with multiple clients and a global server. In this case, the purpose is to train a shared model in a distributed and parallel manner using the local data of the $C$ available clients. Each client will have a different bias because of the conditions of its local environment. Likewise, its data stream may change in different ways over time. Therefore, there may occur concept drifts affecting all clients, some of them, or just one. Thus, we can generalize the problem in the following way: Given a time period $[0,t]$, a set of clients $\{1, \dots,C\}$ , and a set of local samples for each client, which we will denote as $S_{0,t}^j = \{(X_0^j,y_0^j), \dots, (X_t^j,y_t^j)\}$, where each $(X_i^j, y_i^j)$ is one data instance from client $j$. $X_i^j$ is the feature vector, $y_i^j$ is the label, and each local dataset $S_{0,t}^j$ has a certain joint probability distribution $P_{t}^j(X,y)$. 
A \emph{local concept drift} occurs at timestamp $t$ for client $j$ if $\exists t, j: P_{t}^j(X,y) \neq P_{t+1}^j(X,y)$.

Nevertheless, note that a local concept drift does not necessarily have a direct impact on the global federated model. It may be the case that a local drift on device $j$ will result in a change in the distribution of $j$, but not in the joint distribution of all clients, $P_{t}^G(X,y)$. In that case, the federated model will not be affected by the local change, and it can be disregarded. Thus, we can define a \emph{global concept drift} as a distribution change at timestamp $t$ in one or more clients $J \subseteq \{1, \dots,C\}$ such that $\exists t: P_{t}^G(X,y) \neq P_{t+1}^G(X,y)$. As opposed to the previous reasoning, detecting a global concept drift implies that at least one client has detected a local concept drift.

Although we define both local concept drift and global concept drift as a change in the data distribution, we should emphasize that we cannot actually know the real distribution of the data based on the different samples we get. What we do is to deduce how distributions look like based on the samples, so in reality we are measuring the changes in the empirical joint probability distributions, $\hat{P}_t^j(X,y)$ and $\hat{P}_t^G(X,y)$.

We already discussed the possible existence of a change in the joint distribution of a client or a small subset of clients that do not affect the global joint distribution. This could happen mainly for two different reasons: That client's data is the only one that is changing, or the rest of the clients have not detected that change yet. If that is the case, while that client is getting data that differs from that of other clients, the global model may perform poorly on it. Therefore, it could be necessary to consider some adaptive strategy to improve the results on that client~\cite{deng2020adaptive}. However, in our case scenario, we assume that the drift is the same and simultaneously occurs for all the clients. Under these assumptions, local and global concept drifts are indistinguishable.

When a global concept drift happens, the model will probably lower its performance. Hence, as we will illustrate in Section~\ref{sec:drift}, we require to extend the FedAvg algorithm by giving it the ability to detect these global changes and adapt to them on its own.

\section{\ourMethod{} (\ourAcronym{})}
\label{sec:drift}

Algorithms~\ref{alg:cda-fedavg-server}~and~\ref{alg:cda-fedavg-client} detail our proposal. Algorithm~\ref{alg:cda-fedavg-server} shows the pseudocode of the method on the server side, whilst Algorithm~\ref{alg:cda-fedavg-client} exposes the client side. Unlike FedAvg, our approach is asynchronous, so there is no predefined sequence in the order of events and communications between the server and each of the participants. Thanks to concept drift detection and adaptation, each device has enough autonomy to decide \emph{when} to train and \emph{what} data to use for that purpose, so that the server will simply orchestrate the process.

As we can see in Algorithm~\ref{alg:cda-fedavg-server}, the server starts by initializing the global model and communicating it to all the participants (lines~1--2). Then, it will periodically check if there has been any local update on one or more devices (lines~4--5), which will involve performing a global aggregation in order to update the global model too (line~6).  Each time the model is globally consensuated, the server will have to broadcast it to all the clients so that they always have the latest version of the model (line~7). 

\SetKwInOut{kwInput}{Input} 
\SetKwInOut{kwOutput}{Output}
\begin{algorithm}[!htb] 
            \algorithmfootnote{* Given the continual nature, we leave the choice of the stop criteria as a matter of implementation.}
            \kwInput{List of participant clients $\pazocal{C} = \{1,2, \dots, C\}$} 
            \kwOutput{Global model $\mathbf{w}^G$.}
		    Initialize $\mathbf{w}_0^G$ \\
		    \texttt{broadcast}($\mathbf{w}_{0}^G, \pazocal{C}$) \tcp*{Send the model to all clients} 
		    \While{true\textnormal{*}}{  
		        Listen for client updates $\forall j \in \pazocal{C}$ \\
		        \If{$\exists j \in \pazocal{C} :$ new update $\mathbf{w}^j_{t}$ is received}{
    				$\mathbf{w}_{t}^G \leftarrow \sum_{j=1}^{C} \frac{n^j_t}{N}\mathbf{w}^j_{t}$ \tcp*{Averaging aggregation}
        			\texttt{broadcast}($\mathbf{w}_{t}^G, \pazocal{C}$)  
        		}
			}
	\caption{\ourMethod{}, server side.\label{alg:cda-fedavg-server}}
\end{algorithm}

Notice that, using this framework, it is possible for one or more clients to send updates at any time, giving room to different participation rates among them. 
Hence, the global model could be better fitted to the particularities of the most active participants. In order to prevent overfitting of the model to some clients, it could be interesting to consider bounds on the participation rates to control the number of updates per client.

Clients are responsible for learning the task locally, managing, if necessary, the concept drift. 
Research on learning under concept drift is usually divided into three main stages~\cite{lu2018}: (1) drift detection (whether or not a drift occurs), (2) drift understanding (exactly when, how, where and why it occurs) and (3) drift adaptation (response to that drift). Our method focuses on global drift detection and adaptation. To that end, each client (Algorithm~\ref{alg:cda-fedavg-client}) will continually acquire new data from its environment. This data will be processed to identify new concepts (drift detection) and learn from them (drift adaptation). The issue of drift understanding is beyond the scope of this article. We are just concerned about detecting a difference between two timestamps, without analyzing the drift nature, or what causes it.

\SetKwInOut{kwInput}{Input} 
\SetKwInOut{kwOutput}{Output}
\begin{algorithm}[!htb]
            \algorithmfootnote{* Given the continual nature, we leave the choice of the stop criteria as a matter of implementation.}
            \kwInput{Minimum amount of data to train $L$, local minibatch size $B$, number of rounds $R$ per change, number of local epochs $E$ per round, learning rate $\eta$, sensitivity to change $\lambda$, padding $\Delta$, and maximum size $N_{max}$ for the sliding window.} 
            \kwOutput{None.} 
		    $\pazocal{Q} \leftarrow \{\varnothing\}$ \tcp*{Initialize the sliding window (short-term memory)}
		    $\pazocal{L} \leftarrow \{\varnothing\}$ \tcp*{Initialize the long-term memory} 
		    $\pazocal{L} \leftarrow $\texttt{driftAdaptation}($\pazocal{L},L,B,R,E,\eta$)  \tcp*{Learn first concept (Algorithm~\ref{alg:drift-adaptation})} 
			\While{true\textnormal{*}}{
			    \If{new data instance, $X_i$, is observed}{
			        $[\hat{y}_i, q_i] \leftarrow \texttt{predict}(\mathbf{w}^G_t, X_i)$ \tcp*{Classify the pattern} 
			        $\pazocal{Q} \leftarrow \pazocal{Q} \cup q_{i}$ \tcp*{Add the confidence into $\pazocal{Q}$} 
    			    \If{$|\pazocal{Q}| >= N_{max}$}{
    				    $\pazocal{Q} \leftarrow \pazocal{Q} \setminus \{q_1\}$  \tcp*{Remove the oldest element in $\pazocal{Q}$}
        			}
    		        $r \leftarrow \texttt{random}(0,1)$ \tcp*{ Generate random number in the interval [0,1]} 
		            \If{$e^{-2q_i} \geq r$}{ 
    			        $d \leftarrow$ \texttt{driftDetection}($\pazocal{Q},\lambda,\Delta,N_{max}$) \tcp*{ Check for drift (Algorithm~\ref{alg:drift-detection})} 
    			        \If{$d$ is true}{
    			            $\pazocal{Q} \leftarrow \{\varnothing\}$ \\ 
    			            $\pazocal{L} \leftarrow $\texttt{driftAdaptation}($\pazocal{L},L,B,R,E,\eta$) \tcp*{Update (Alg.~\ref{alg:drift-adaptation})} 
    			        }
			        }
				}
			}
			\caption{\ourMethod{}, client side.\label{alg:cda-fedavg-client}}
\end{algorithm}

Regarding the detection and adaptation, we propose that each client handles two different data storages: a \emph{short-term memory} and a \emph{long-term memory}. The short-term memory, $\pazocal{Q}$, is used to store the data instances the device has acquired in the last time interval. This recent data is kept for a limited amount of time and is processed to check whether a drift occurs. The long-term memory, $\pazocal{L}$, will store data samples from each of the concepts seen so far. This information will be kept for a long time and will be used to train and retrain the model locally, so that all concepts are learned.

Basically, each client operates as follows (Algorithm~\ref{alg:cda-fedavg-client}): At the beginning of the process, both short-term and long-term memories are empty (lines~1--2), and the model has never been trained locally. Thus, the first data acquired by each client will automatically belong to the initial concept. This data will be stored in the long-term memory and used to perform the first local update (line~3). After that, each client continues to acquire new data, saving it in the short-term memory and processing it to check whether a drift is detected (lines~5--13). Only when the drift detection algorithm confirms the drift (line~14), new data related to the new concept will be stored in the long-term memory and further training rounds will be carried out (line~16). In the following two subsections, we will discuss in more detail the two fundamental parts of our method on the client side: drift detection~(Section~\ref{sec:drift-detection}) and drift adaptation~(Section~\ref{sec:drift-adaptation}).

\subsection{Drift Detection}
\label{sec:drift-detection}
Drift detection encompasses the range of procedures and techniques that recognize and quantify concept drift via identifying change points or change intervals in the underlying distribution of data. In our context, detecting concept drift implies that the federated model is no longer a good predictor for all the clients and must be adapted.
Drift detection algorithms generally fall into three categories~\cite{lu2018}: (1) error rate-based methods, (2) data distribution-based methods and (3) multiple hypothesis test methods. The algorithms of the first group focus on tracking changes in the online error rate of the model. The second class uses a distance function or metric to quantify the dissimilarity between the distribution of historical data and that of new data. The third category combines techniques from the two previous ones in several ways. 
In this work we decided to use a data distribution-based algorithm. Our choice is based on being able to detect virtual concept drift without needing pattern labels as input.

We propose a CUSUM-type detection method based on \textit{beta} distribution~\cite{baron1999convergence}, which is inspired in the original work of Haque et al.~\cite{haque2016}.
It is executed on each of the devices as soon as a new data instance $(X_i, y_i)$ is available (line~5, Alg.~\ref{alg:cda-fedavg-client}). As this method does not require instance labels to detect the drift, we actually only need to get the new feature vector $X_i$. 
For each new $X_i$, we will estimate a metric that will help us to quantify the dissimilarity between the old and the new data distributions. Any of the metrics that have been proposed in the literature can be used~\cite{lu2018}. We chose the confidence of the classifier as metric because our experimental evaluation is carried out on a classification task (Section~\ref{sec:exp-results}). We define the \emph{confidence} of the model on a sample $(X, y)$ as the classifier maximal conditional posterior probability. That means that we use the federated model to predict the class of the pattern $X$, and we get the maximal probability $\textnormal{P}(c_k | X)$, where $c_k$ is one of the $M$ possible classes $\pazocal{Y} = \{c_1, c_2, \dots, c_M\}$. This corresponds to line~6 in Algorithm~\ref{alg:cda-fedavg-client}.
The confidence $q_i$ of the current model on the new data instance $X_i$ is then stored in the short-term memory of the client  (line~7, Alg.~\ref{alg:cda-fedavg-client}). This short-term memory is a sliding window $\pazocal{Q}$ of length $N = |\pazocal{Q}|$.

We aim to identify changes in the distribution of the confidences stored in $\pazocal{Q}$. In the original method proposed in~\cite{haque2016}, the authors do not use a sliding window, but a dynamic window. They suggest reinitializing this dynamic window each time a concept drift is detected, but they do not establish any limits on its size. This is quite unrealistic, because it could grow to infinity if no drifts are detected. Therefore, we use a sliding window instead, and we set its maximum size $N_{max}$. Once this maximum size is reached, adding a new element to $\pazocal{Q}$ implies deleting the oldest one (lines~8--10, Alg.~\ref{alg:cda-fedavg-client}). Following this approach, we are able to detect sudden, recurring and gradual drifts, and even incremental ones if the sliding window is big enough to cover data sufficiently distinct. Therefore, the value of $N_{max}$ will be strongly dependent on the task to be solved. In our experiments (Section~\ref{sec:exp-results}), we are interested in sudden concept drifts, and we set $N_{max} = 1000$.

The core of our detection method is detailed in Algorithm~\ref{alg:drift-detection}, which is called by Algorithm~\ref{alg:cda-fedavg-client} in line~13. Algorithm~\ref{alg:drift-detection} can be a bottleneck in the system if we have to run it after inserting each confidence value in $\pazocal{Q}$. Consequently, we limit the number of executions, so that Algorithm~\ref{alg:drift-detection} will be run with a probability of $e^{-2q_i}$, for any confidence value $q_i$ (line~12 in Algorithm~\ref{alg:cda-fedavg-client}). Hence, the higher the confidence, the lower the likelihood of executing the drift detection, and \textit{vice versa}.

During drift detection (Algorithm~\ref{alg:drift-detection}), $\pazocal{Q}$ is split into two sub-windows for every pattern $k$ between $\Delta$ and $N - \Delta$, where $N$ is the length of $\pazocal{Q}$ (lines~4--6). 
Let $\pazocal{Q}_a$ and $\pazocal{Q}_b$ be the two sub-windows, where $\pazocal{Q}_a$ contains the most recent confidences. Each sub-window is required to contain at least $\Delta$ elements to maintain the statistical properties of a distribution. When a concept drift occurs, it is expected that confidence scores will decrease. Therefore, we only need to detect changes in the negative direction. Namely, if $m_a$ and $m_b$ are the mean values of the confidences in $\pazocal{Q}_a$ and $\pazocal{Q}_b$ respectively, a change point is searched only if $m_a \leq (1-\lambda) \times m_b$, where $\lambda$ is the sensitivity to change (line~7). Same as in~\cite{haque2016}, we use $\lambda = 0.05$ and $\Delta = 100$ in our experiments, which are also widely used in the literature.

\SetKwInOut{kwInput}{Input} 
\SetKwInOut{kwOutput}{Output}
\begin{algorithm}[!htb]
            \kwInput{Sliding window $\pazocal{Q}$, sensitivity to change $\lambda$, and padding $\Delta$ and maximum size $N_{max}$ for the sliding window.} 
            \kwOutput{Boolean indicating whether a drift is detected or not.}
            $s_f \leftarrow 0$ \\
            $T_h \leftarrow - \log(\lambda)$ \\
            $N \leftarrow |\pazocal{Q}|$ \\
			\For{$k \leftarrow \Delta$ \KwTo $N - \Delta$}{
			    $m_b \leftarrow \texttt{mean}(q_1  : \ q_k \ \in \pazocal{Q})$ \\
			    $m_a \leftarrow \texttt{mean}(q_{k+1}  : \ q_N \ \in \pazocal{Q})$ \\
			    \If{$m_a \leq (1- \lambda) \cdot m_b$}{
			        $s_k \leftarrow 0$ \\
			        $[\hat{\alpha}_b, \hat{\beta}_b] \leftarrow \texttt{estimateParams}(q_1  : \ q_k)$  \tcp*{Get parameters of beta distribution}
			        $[\hat{\alpha}_a, \hat{\beta}_a] \leftarrow \texttt{estimateParams}(q_{k+1}  : \ q_N)$ \\
					\For{$i \leftarrow k+1$ \KwTo $N$}{
					    $s_k \leftarrow s_k + \log\left(\frac{f\left(q_i \ | \ \hat{\alpha}_a, \ \hat{\beta}_a\right)}{f\left(q_i  \ | \ \hat{\alpha}_b, \ \hat{\beta}_b\right)}\right)$ \\
					}
					$s_f \leftarrow \texttt{max}(s_f, s_k)$ \\
				}
			}
			\uIf{$s_f > T_h$}{
			    \Return \textit{true}
			}\Else{
			    \Return \textit{false}
			}
	\caption{Drift detection method (\texttt{driftDetection}).\label{alg:drift-detection}}
\end{algorithm}

The confidence values in each sub-window ($\pazocal{Q}_a$ and $\pazocal{Q}_b$) are expected to follow two different \textit{beta} distributions. However, the actual parameters for each one are unknown. Algorithm~\ref{alg:drift-detection} estimates these parameters at lines~9 and 10, given the mean and the variance of each sub-window, by using the \emph{method of moments}~\cite{bowman2014estimation}. Next, the sum of the log likelihood ratios $s_k$ is calculated in the inner loop between lines~11 and 13, where $f\left(q_i \ | \ \hat{\alpha},  \hat{\beta}\right)$ is the probability density function (PDF) of the \textit{beta} distribution, having estimated parameters $\left(\hat{\alpha}, \hat{\beta}\right)$, applied on the confidence $q_i \in \pazocal{Q}$. 
This PDF describes the relative likelihood for a random variable, in this case $q_i$, to take on a given value, and it is defined as:
\begin{equation*}
    f\left(q_i \ | \ \alpha,  \beta\right) = \begin{cases}
    \frac{q_i^{\alpha-1}(1-q_i)^{\beta-1}}{B\left(\alpha, \beta\right)}, & \text{if } 0 < q_i< 1 \\
    0, & \text{otherwise,} 
  \end{cases}
\end{equation*}
where
\begin{equation*}
     B\left(\alpha,  \beta\right) = \int_{0}^{1} q_i^{\alpha-1}(1-q_i)^{\beta-1} dq_i.
\end{equation*}

The variable $s_k$ is a dissimilarity score for each iteration $k$ of the outer loop between lines~4 and 16. The larger the difference between the PDFs in $\pazocal{Q}_a$ and $\pazocal{Q}_b$, the higher the value of $s_k$~(line~12).
Let $k_{max}$ is the value of $k$ for which the algorithm calculated the maximum $s_k$ value where $\Delta \leq k \leq N - \Delta$. Finally, a change is detected at point $k_{max}$ if $s_{k_{max}} \equiv s_f$ is greater than a prefixed threshold $T_h$~(line~17). As in the original work, we use $T_h = - \log(\lambda)$.

In case a drift is detected, the sliding window $\pazocal{Q}$ is reinitialized and the drift adaptation method is called~(lines~14--17 in Algorithm~\ref{alg:cda-fedavg-client}). We will discuss the adaptation strategy in the next subsection.

\subsection{Drift Adaptation}
\label{sec:drift-adaptation}
Once a concept drift is detected, the model should continue to be trained being aware of that drift. Otherwise, catastrophic forgetting may occur. When dealing with neural networks, as is the case here, the most common methods to avoid this are (1) regularization methods, (2) rehearsal approaches, and (3) generative replay~\cite{lesort2020continual}. In continual learning, regularization consists of protecting the important weights of previous concepts from modification. Rehearsal or replay approaches are based on saving raw samples as a memory of past tasks. Generative replay techniques train generative models on the data distribution, thus being able to sample data from past experiences when learning on the new one.

Rehearsal methods have been shown to provide the best results, as they ensure that the memories do not degrade over time~\cite{van2019three}. When working in a cloud-centric setting, the main drawback of these techniques is the need for a separate memory of raw data. This is a vanilla way of saving knowledge that does not respect data privacy and may involve high storage requirements. Nonetheless, in the federated (and therefore distributed) setting, this is not a problem since small pieces of old memories can be stored locally by each of the clients. Thus, we propose to use a simple rehearsal method, which is detailed in Algorithm~\ref{alg:drift-adaptation}.

Algorithm~\ref{alg:drift-adaptation} is called if a new drift is detected (using the detection method from Section~\ref{sec:drift-detection}). This method is responsible for managing the long-term memory of the client and using this data to update the model locally. Formally, the long-term memory of a client $j$ at time $t$ is a dataset $\pazocal{L}^j = \{\pazocal{L}_0^j \cup \pazocal{L}_m^j \cup ... \cup \pazocal{L}_l^j\}$, which gathers representative data of each concept $\{\kappa_0^j, \kappa_m^j, \dots, \kappa_l^j\}$ detected so far by $j$, where $0, m, \dots, l$ are the timestamps where each drift was detected, being $0 < m < \dots < l \leq t$. 

\SetKwInOut{kwInput}{Input} 
\SetKwInOut{kwOutput}{Output}
\begin{algorithm}[!htb] 
            \kwInput{Long-term memory $\pazocal{L}$, minimum amount of data to train $L$, minibatch size $B$, number of rounds $R$ per change, number of local epochs $E$ per round, and learning rate $\eta$.} 
            \kwOutput{Updated long-term memory $\pazocal{L}$.} 
            $\pazocal{L}_{new} \leftarrow \texttt{collectData}()$ \tcp*{Collect enough new data on the new concept} 
                $\pazocal{L} \leftarrow \pazocal{L} \cup \pazocal{L}_{new}$ \tcp*{Expand the long-term memory with the new data} 
            \tcc{Perform local training just in this client} 
            \For{$r \leftarrow 1$ \KwTo $R$}{
		        Split $\pazocal{L}$ into a set $\pazocal{B}$ of batches of size $B$ \\
    			\For{$e \leftarrow 1$ \KwTo $E$}{
    			    \For{\textnormal{batch} $b \in \pazocal{B}$}{
    			        $\mathbf{w}_{t+1} \leftarrow \mathbf{w}_{t}^G - \eta \nabla l (\mathbf{w}_{t}^G; b)$ \tcp*{ $\nabla l$ is the gradient of $l$ on $b$} 
    				}
    			}
    			Send $\mathbf{w}_{t+1}$ to server to perform global aggregation
			}
	\caption{Drift adaptation method (\texttt{driftAdaptation}). \label{alg:drift-adaptation}}
\end{algorithm}

At the beginning of Algorithm~\ref{alg:drift-adaptation}, the long term memory does not contain any data instance related to the new concept $\kappa_{new}^j$. Therefore, the first thing we do is to collect enough data $\pazocal{L}_{new}^j \subset \pazocal{L}^j$ belonging to $\kappa_{new}^j$ (line~1). All data that was obtained after the cut-off point $k_{max}$ where the last drift was detected (Algorithm~\ref{alg:drift-detection}) can be immediately included in $\pazocal{L}_{new}^j$.
In continual classification tasks, the data from each of the classes can be collected in any order. Thus, to guarantee a minimally balanced rehearsal dataset with the representation of all classes, we define a heuristic rule to control the data which is saved in the long-term memory. We establish a minimum amount of data, $L$, so that there must be at least $\frac{L}{2M}$ examples from each class $c \in \pazocal{Y}$ in the long-term memory representing each concept, where $M$ is the number of possible classes. Formally, client $j$ keeps collecting data until the following condition is met:
\[
    \forall c \in \pazocal{Y} :  |\{(X,y) \in \pazocal{L}_l^j : y=c \}| \geq \frac{L}{2 M}.
\]
In our experiments, we have set $L = 1400$, which is quite a significant amount of data, but it ensures that the memories do not degrade over time. We also assume that we can keep data for an unlimited number of concepts.

Once enough data on the new concept is gathered, local training can be carried out (lines~3--11). Local training is conducted during a limited number of rounds $R$. In our experiments we have used $R = 5$, although this value will depend on the problem.
Between rounds, each local update $\mathbf{w}_{t+1}^j$ is communicated to the server, where the global aggregation is done (line~15).

\subsection{Analysis of computational complexity} 
We will now give some insights into the complexity of \ourAcronym{}, compared to regular FedAvg. We will analyze three different aspects: (1) time, (2) communications, and (3) memory requirements.

Before comparing the costs of both methods, we need to set some assumptions to get upper bounds on the number of operations required. 
\begin{enumerate}
    \item Firstly, we assume that our loss functions $l_j$'s, which are the same for all clients, are H-smooth. This means that they are differentiable, so we can calculate their gradients, and at the same time their gradients are H-Lipschitz functions \cite{armijo1966minimization}, which can be expressed as follows:
    \[ \exists H \in \mathbb{R}^+ : \ \ \mid \mid \nabla l_j(u;b) - \nabla l_j(v;b) \mid \mid \ \leqslant H \mid \mid u - v \mid \mid \ \ \ \forall \ u, v \in \pazocal{W}, u \neq v,  \]
    where $\pazocal{W}$ denotes the set of all possible weights, and $b$ is the selected batch of data. This equation has a geometrical interpretation:
    \[ \mid \mid \nabla l_j(u;b) - \nabla l_j(v;b) \mid \mid \ \leqslant H \mid \mid u - v \mid \mid \ \Leftrightarrow \ \frac{ \mid \mid \nabla l_j(u;b) - \nabla l_j(v;b) \mid \mid }{ \mid \mid u - v \mid \mid} \leqslant H. \]
    This last equation, when $u$ tends to $v$, is the definition of the derivative function, so this inequality says the derivative of $\nabla l_j$ is upperly bounded by a constant value $H$.
    
    \item Secondly, we also assume that the expectation of the difference between the variations of the local loss functions for each client $l_j(u;b)$ and the variation of the global loss function $\ell(u)$ is upperly bounded. We define this global loss function as the expected error in all clients when using the weights $u$:
    \[ \ell(u) = \underset{b_j \thicksim P_j^t}{\mathbb{E}} \left[ \sum_{j=1}^C l_j(u;b_j) \right].\]
    Given this global loss function, our assumption of proximity could be expressed as follows:
    \[ \exists \sigma \in \mathbb{R}^+ : \mathbb{E} \left[ \ \mid \mid \nabla_{u} l_j(u;b) - \nabla \ell(u) \mid \mid \ \right] \leqslant \sigma^2 \ \ \ \ \forall j \in \{1,\ldots,C\}. \] 
    
    \item Lastly, we assume the global loss function we try to minimize is convex: 
    \[ \ell(\mu u + (1-\mu)v) \leq \mu \ell(u) + (1-\mu)\ell(v) \ \ \ \  \forall \ u, v \in \pazocal{W}, \forall \mu \in [0,1]. \]
    This guarantees the global loss minimization problem has a unique solution, and there are no local minimums except for the global minimum.
    
\end{enumerate}

Taking into account the above assumptions, we can give an upper bound for the time complexity (number of elementary operations) of both FedAvg and \ourAcronym{}. Kairouz et al.\cite{kairouz2019advances} states that the complexity of FedAvg in terms of the number of training rounds $R$, the number of local epochs $E$, and the number of clients selected to train on each round $m$, is:
\begin{equation}\label{FedAvgCalculations}
    \pazocal{O}\left( \frac{H}{R^2} + \frac{\sigma}{\sqrt{REm}} \right).
\end{equation}
In our case, under the same assumptions, we carry out the same operations, but we also check whether or not concept drift occurs using Algorithm~\ref{alg:drift-detection}. All of those further calculations depend linearly on the maximum size of the sliding window, $N_{max}$, which depends linearly on the minimum number of data needed, $2 \Delta$, so our upper bound is shortly bigger:
\begin{equation}\label{DriftCalculations}
    \pazocal{O}\left( \frac{H}{R^2} + \frac{\sigma}{\sqrt{REm}} + \Delta \right).
\end{equation}
However, note that in \ourAcronym{} the number of training rounds $R$ is typically fewer than in FedAvg since, after detecting and adapting to a drift, we stop training until a new concept is detected.

In the case of the second aspect we want to consider, communications, \ourAcronym{} is expected to make fewer exchanges of data than FedAvg. In the original method, the communications are bounded by
\begin{equation}\label{FedAvgCommunications}
    \pazocal{O} (Rm).
\end{equation} 
Nevertheless, in our case, if a drift is detected, the next data communication takes place after adapting the local model to that drift. If we set $\nu$ as the number of times a drift is detected divided by the total number of times Algorithm~\ref{alg:drift-detection} is executed, then the number of communications our method performs is
\begin{equation}
    \pazocal{O} \left( Rm(1-\nu) + Rm \frac{M\nu}{L} \right) = \pazocal{O} \left(  Rm (1- \nu) \right),
\end{equation}
where $M$ is the number of possible classes, and $L$ is a minimum amount of data needed in the long-term memory from each concept. Therefore, if there are no drifts $(\nu = 0)$, and assuming the number or rounds $R$ is the same as in FedAvg (although \ourAcronym{} typically uses a lower $R$), we get the same cost of Equation~\eqref{FedAvgCommunications}. However, in any other case, the communication cost of \ourAcronym{} gets reduced proportionally to the total number of communications expected. In order to compare both cost differences, we can express $\nu$ in terms of $\Delta$. Each time a drift is detected, we set the short-term memory $\pazocal{Q} = \varnothing$. To detect subsequent drifts, the sliding window needs to be at least of size $2 \Delta$, which implies $\nu \leqslant \frac{1}{2 \Delta}$. Thus, we can rewrite
\begin{equation}\label{DriftCommunications}
    \pazocal{O} \left( Rm(1 - \nu) \right) = \pazocal{O} \left( Rm - Rm\nu \right) \leqslant \pazocal{O} \left( Rm - \frac{Rm}{2 \Delta} \right).
\end{equation}

Finally, regarding the last aspect we want to compare, memory costs, we must be aware of the fact that, in FedAvg, each client needs to storage all the data obtained locally. Nevertheless, in our case, each client keeps data just until there are enough samples of the current concept. After that, no more storage is needed until detecting a new drift.

\subsection{Benefits of Explicit Drift Detection: Saving~Resources~on~Devices}
\ourAcronym{} not only allows continual learning that avoids catastrophic forgetting, but also answers two fundamental questions that are often overlooked in the literature on federated learning: \emph{what} to learn and \emph{when} to learn it.

As we already mentioned, in a naive implementation of FedAvg there is no given criterion on what data to store and what to discard. By default, unless there are task-dependent or implementation restrictions, devices will save all the data they collect. Besides, FedAvg does not set any limits on the number of training rounds to be performed, which could be infinite. It is true that only a random subset of devices is selected in each round, so they are not training all the time. In addition, it is possible to limit training even further based on conditions such as whether the device has access to WiFi or is connected to a power supply (this is specially intended for smartphones)~\cite{konevcny2016federated,hard2018federated}. However, these criteria do not take into account whether there is a real need for further training. Finally, training constantly and without proper criteria also implies an increase in the required bandwidth. In fact, in many works in the literature of federated learning authors state that the main bottleneck is the communication cost~\cite{mcmahan2016federated,yoon2020federated}.

Following our approach, it is possible to determine which data is relevant and which is not anymore thanks to the integration of the concept drift detector and a long-term memory. We do not quantify the difference of storage explicitly, but in the FedAvg approach, storage grows linearly with the number of training rounds $R$, whereas in our case it stops at some point, when we reach a sufficient number of samples. Besides, it can be assumed that it is only necessary to continue training when a new drift is detected, so that the model can be adapted to it. As shown in Equations \eqref{FedAvgCalculations} and \eqref{DriftCalculations}, our approach needs to perform more calculations for the same number of training rounds, at most $\pazocal{O}(\Delta)$. However, if there are no drifts, we can pause  the training process at some point, reducing this cost. In addition, Equations \eqref{FedAvgCommunications} and \eqref{DriftCommunications} show that we save at least $\pazocal{O}\left(\frac{Rm}{2 \Delta}\right)$ communications.
Thus, we prove it is possible to significantly reduce the number of communications between clients and server, as well as the computational burden on clients.

\section{Experimental Results}
\label{sec:exp-results}
\begin{figure}[htb]
	\centering
	\includegraphics[width=0.5\textwidth,trim={0 2 0 2},clip]{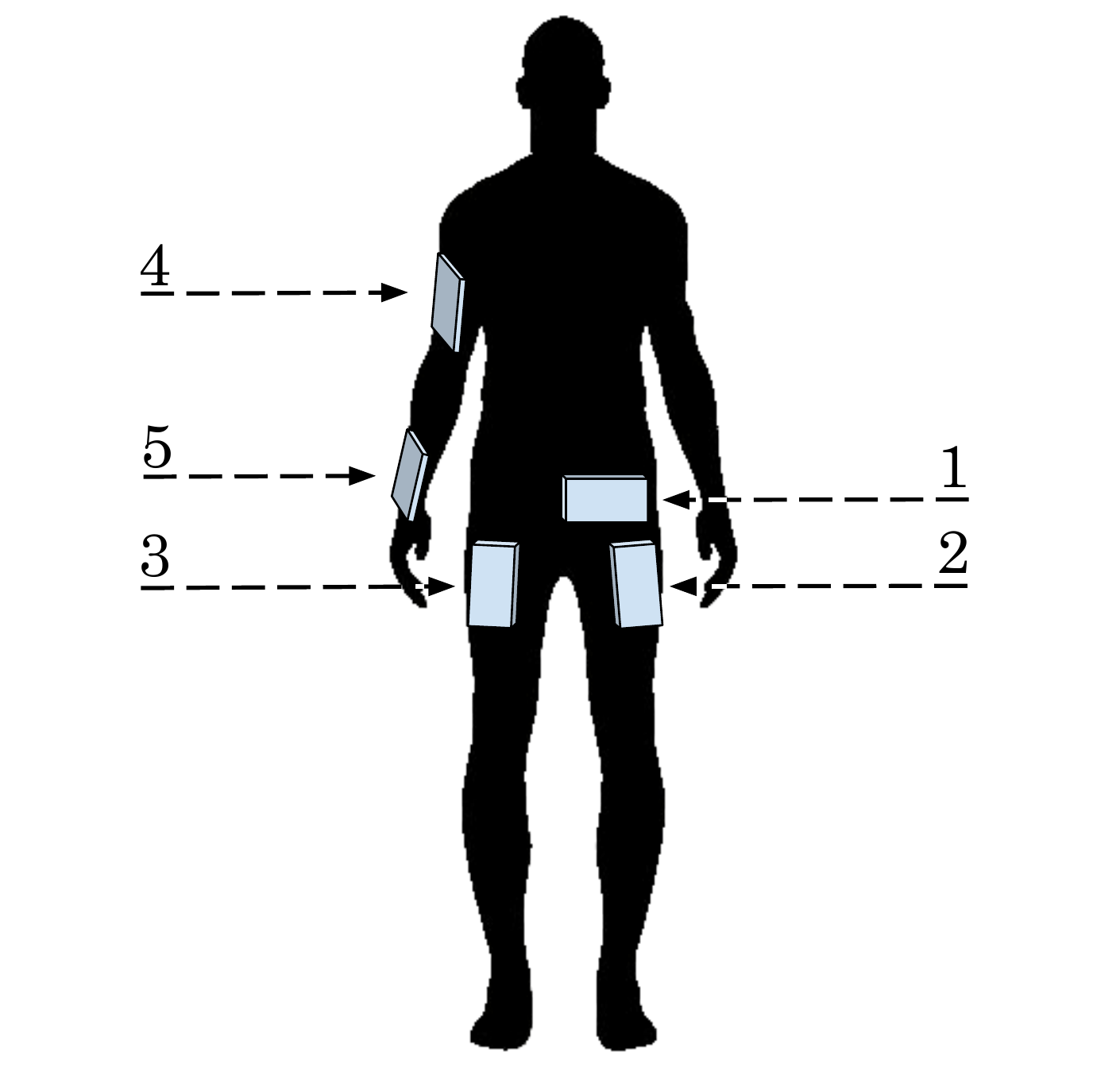}
	\caption{Overview of the phone positions on a participant.} 
	\label{fig:phone-positions}
\end{figure}

To test \ourAcronym{}, we have chosen a complex real-world task: Human Activity Recognition (HAR) on smartphones. More specifically, we have used the dataset from Shoaib et al.~\cite{shoaib2014fusion}, which includes data of seven physical activities: walking, going upstairs, going downstairs, sitting, standing, jogging, and biking. This dataset is characterized by having, for each activity, data recorded with the smartphone placed in 5 different locations (Figure~\ref{fig:phone-positions}): (1)~on the belt position using a belt clip, (2)~in the left jeans pocket, (3)~in the right jeans pocket, (4)~on the right upper arm using a sports armband, and (5)~on the right wrist. This last element is essential for us because, as we will see later, it will allow us to analyze the performance of \ourAcronym{} in response to changes.

The dataset gathers data from 10 different people. Everyone performed each of the 7 activities for about 20 minutes (4 minutes for each phone position). The ten participants were male, aged between 25 and 30 years. All experiments were conducted indoors in the same building, except for the biking ones, which took place outside.
The smartphones were oriented vertically for the upper arm, wrist, and two pockets, and horizontally for the belt position. The data that was collected includes the readings from the inertial sensors of the smartphone, comprising the tri-axis accelerometer, gyroscope, and magnetometer. The data acquisition rate was limited to 50 samples per second because it is enough to recognize human physical activities.

We pose the problem as a multiclass classification task where the goal is to correctly predict which of the 7 activities is being performed by the user. The information about the position of the smartphone will be used later to simulate concept drifts.
In order to create a federated learning setting, we choose 9 of the 10 phone users as clients, in which both FedAvg and \ourAcronym{} will be running. Data from the remaining participant is used for testing. In the experiments that we will show below, we performed cross-validation. Thus, in the 9--1 split between training and testing, the test user was permuted a total of 10 times, so that each experiment was repeated 10 times. Since there are relatively few devices, at each federated round of each execution, all 9 clients participated in the local training, instead of selecting a random subset.

We split the original raw data into windows of 124 samples, which is equivalent to about $2.5$ seconds. We decided to use just the accelerometer and gyroscope signals. Thus, the input shape is a 6-dimensional time window consisting of 124 values for each of the axes of the accelerometer ($a_x$, $a_y$, $a_z$) and the gyroscope ($\omega_x$, $\omega_y$, $\omega_z$). Each user has a total of 5000 time windows or instances, 1000 for each phone location. In total, in each execution we used 45000 samples for training (distributed among the 9 clients) and 5000 for testing.

For the model, we used a Convolutional Neural Network (CNN), given the performance this type of network has shown for inertial signal processing~\cite{casado2020walking,tong2021cnn}. However, there is nothing preventing \ourAcronym{} algorithm from being applied using a different type of network, like any other feed-forward architecture; recurrent neural networks (RNNs) such as long short-term memory (LSTM); or even a hybrid approach. The architecture we propose has 6 input channels and consists of two 1D convolutional layers, one max-pooling layer, one flattening layer, two dropout layers and two fully connected layers. The total number of learnable parameters is 764399. Table~\ref{tab:cnn} shows the details of the architecture. 
A dropout rate of 0.2 was used in both dropout layers. We used a batch size of 100 instances. On each local training round, 10 epochs were performed by each client. For simplicity in presenting the results, we assume all clients begin to acquire data at the same time and at the same frequency.
\begin{table}[htb]
\centering
\caption{Details of the CNN architecture used in the experiments.}
\label{tab:cnn}
\begin{tabular}{l|ccccc}
\hline\noalign{\smallskip}
Layer name & Kernel size & \# kernels & Stride & Feature map. & \# params  \\
\noalign{\smallskip}\hline\noalign{\smallskip}
conv1                          & 1x10         & 100          & 1      & 115x100     & 6100          \\
conv2                          & 1x10         & 100          & 1      & 106x100     & 100100        \\
max\_pool                      & 1x2         & -           & 1      &  53x100     & 0               \\
dropout1                       & -           & -           & -      &  53x100     & 0               \\
flattening                     & -           & -           & -      &  1x5300     & 0               \\
fully\_con1                    & -           & -           & -      &  1x124      & 657324          \\
dropout2                       & -           & -           & -      &  1x124      & 0               \\
fully\_con2                    & -           & -           & -      &  1x7        & 875             \\
\noalign{\smallskip}\hline
\end{tabular}%
\end{table}

\begin{figure}[htbp]
	\centering
	\includegraphics[width=0.9\textwidth,trim={0 10 15 50},clip]{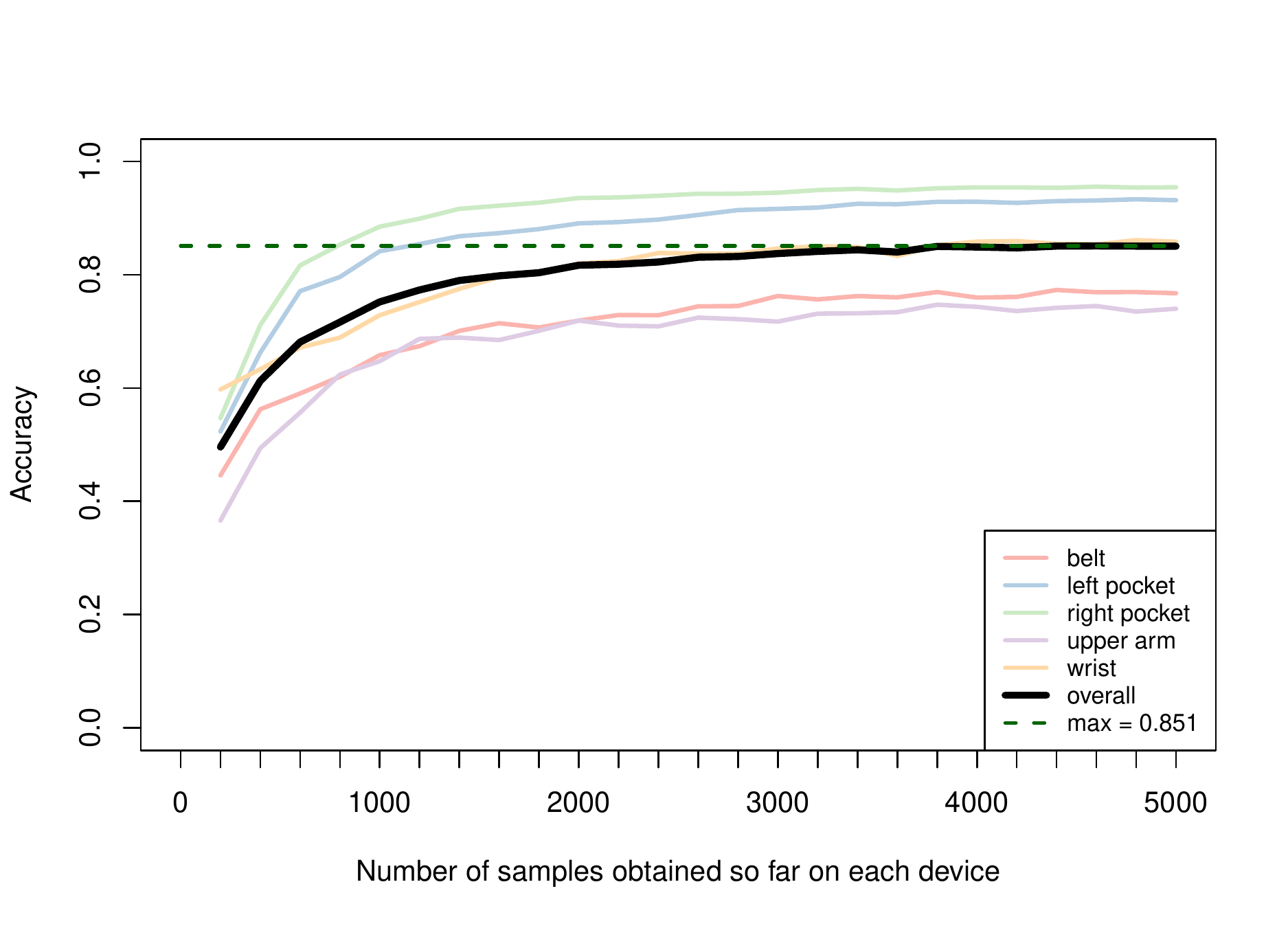}
	\caption{Average test results for standard FedAvg in an IID setting. The thick black line is the overall test accuracy.}
	\label{fig:fedavg-random}
\end{figure}

To provide a baseline, we first carried out federated learning under the unrealistic assumption that data is acquired in an identically distributed manner over time and for all users. Thus, we simply shuffled all the data of each user (5000 samples) randomly, without taking into account the position of the phone, thus forcing a stationary situation (IID in the temporal domain). We applied regular FedAvg algorithm during a total of 25 rounds of local training and server aggregation. We repeated the experiment 10 times, each time leaving a different client for testing. On average, at the end of the process, we achieve over 85\% accuracy on test. Figure~\ref{fig:fedavg-random} shows the average evolution over time. This plot does not show any of the individual executions, but the average values of the 10 executions. The thick black line is the overall accuracy, whilst each of the thin coloured lines represent how well the model performs when evaluated on the test data specific to each phone position. Table~\ref{tab:fedavg-random} shows the final results, when all clients have collected their 5000 samples, for each phone position and for each of the 10 executions and the average. 

\begin{table}[htbp]
\centering
\caption{Final results for all the executions of standard FedAvg in an IID setting, after all clients have collected 5000 instances.}
\label{tab:fedavg-random}
\resizebox{\textwidth}{!}{%
\begin{tabular}{l|cccccc}
\hline\noalign{\smallskip}
Test set & Belt  & Left pocket & Right pocket & Upper arm & Wrist & Overall \\
\noalign{\smallskip}\hline\noalign{\smallskip}
User 1   & 0.878 & 0.911       & 0.862        & 0.909     & 0.802 & 0.872  \\
User 2   & 0.726 & 0.981       & 0.982        & 0.720     & 0.803 & 0.842  \\
User 3   & 0.891 & 0.951       & 0.898        & 0.741     & 0.898 & 0.876  \\
User 4   & 0.906 & 0.982       & 0.984        & 0.555     & 0.909 & 0.867  \\
User 5   & 0.925 & 0.981       & 0.984        & 0.753     & 0.919 & 0.912  \\
User 6   & 0.672 & 0.981       & 0.970        & 0.709     & 0.824 & 0.831  \\
User 7   & 0.606 & 0.834       & 0.977        & 0.915     & 0.776 & 0.822  \\
User 8   & 0.881 & 0.993       & 0.992        & 0.673     & 0.924 & 0.893  \\
User 9   & 0.445 & 0.992       & 0.989        & 0.635     & 0.819 & 0.776  \\
User 10  & 0.744 & 0.711       & 0.907        & 0.791     & 0.910 & 0.813  \\
Average  & 0.767 & 0.932       & 0.955        & 0.740     & 0.858 & 0.850  \\
\noalign{\smallskip}\hline
\end{tabular}%
}
\end{table}

\begin{figure}[htbp]
	\centering
	\includegraphics[width=0.9\textwidth,trim={0 10 15 50},clip]{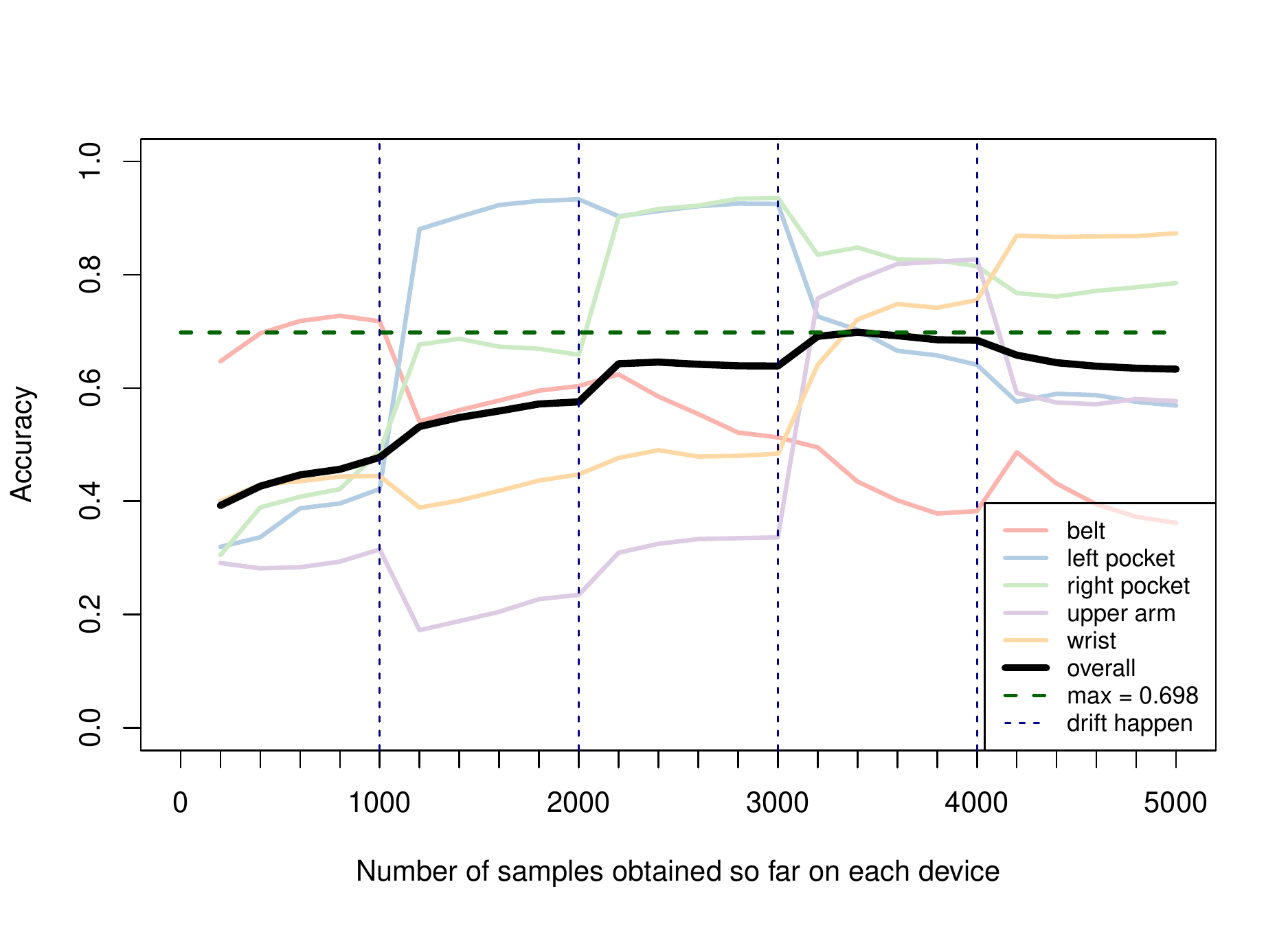}
	\caption{Average test results for FedAvg in a non-IID and nonstationary setting. The vertical dashed lines indicate when a distribution change occurs.}  
	\label{fig:fedavg-drifts}
\end{figure}

Although the results in Figure~\ref{fig:fedavg-random} are quite good, in real life the data is often non-IID and evolves over time. Hence, in our second experiment, we sorted the data of all the users by phone position. In this way, we force a nonstationary distribution that changes a total of 4 times. Each change in the placement of the device implies a change in the underlying distribution of the data, i.e., a concept drift. Data is sorted according to the phone position in the same way for all users: 1\textsuperscript{st}~belt, 2\textsuperscript{nd}~left pocket, 3\textsuperscript{rd}~right pocket, 4\textsuperscript{th}~upper arm, and 5\textsuperscript{th}~wrist. For each location, the corresponding 1000  data observations are randomly sorted. This is not a totally realistic scenario since in real life each user would acquire data in a particular manner, but it is helpful to constrain the changes and check their impact during training. Again, we applied regular FedAvg algorithm with exactly the same configuration as before, training a total of 25 rounds. Figure~\ref{fig:fedavg-drifts} shows the average results of the 10 executions. The vertical dashed lines indicate where a distribution change occurs, which is every 1000 patterns for all the users. It can be seen that, although the general tendency of the model is to improve, it forgets concepts as it learns others. For example, from iteration 1000 it begins to learn about the left pocket position, which causes a drop in accuracy for data related to the previous concept, the belt. The final average accuracy of the model is around 63\%.  Table~\ref{tab:fedavg-drifts} shows the results of the 10 executions when all clients have collected 5000 samples.

\begin{table}[htbp]
\centering
\caption{Final results for all the executions of FedAvg in a non-IID and nonstationary setting, after all clients have collected 5000 instances.}
\label{tab:fedavg-drifts}
\resizebox{\textwidth}{!}{%
\begin{tabular}{l|cccccc}
\hline\noalign{\smallskip}
Test set & Belt  & Left pocket & Right pocket & Upper arm & Wrist & Overall \\
\noalign{\smallskip}\hline\noalign{\smallskip}
User 1   & 0.349 & 0.725       & 0.626        & 0.586     & 0.776 & 0.612  \\
User 2   & 0.358 & 0.598       & 0.735        & 0.633     & 0.773 & 0.619  \\
User 3   & 0.498 & 0.375       & 0.780        & 0.610     & 0.758 & 0.604  \\
User 4   & 0.359 & 0.659       & 0.642        & 0.404     & 0.873 & 0.587  \\
User 5   & 0.349 & 0.666       & 0.807        & 0.706     & 0.978 & 0.701  \\
User 6   & 0.352 & 0.531       & 0.758        & 0.460     & 0.936 & 0.607  \\
User 7   & 0.329 & 0.547       & 0.921        & 0.644     & 0.926 & 0.673  \\
User 8   & 0.375 & 0.531       & 0.873        & 0.525     & 0.934 & 0.648  \\
User 9   & 0.144 & 0.642       & 0.799        & 0.567     & 0.950 & 0.620  \\
User 10  & 0.505 & 0.415       & 0.913        & 0.634     & 0.830 & 0.659  \\
Average  & 0.362 & 0.569       & 0.785        & 0.577     & 0.873 & 0.632  \\
\noalign{\smallskip}\hline
\end{tabular}%
}
\end{table}

\begin{figure}[htbp]
	\centering
	\includegraphics[width=0.9\textwidth,trim={0 10 15 50},clip]{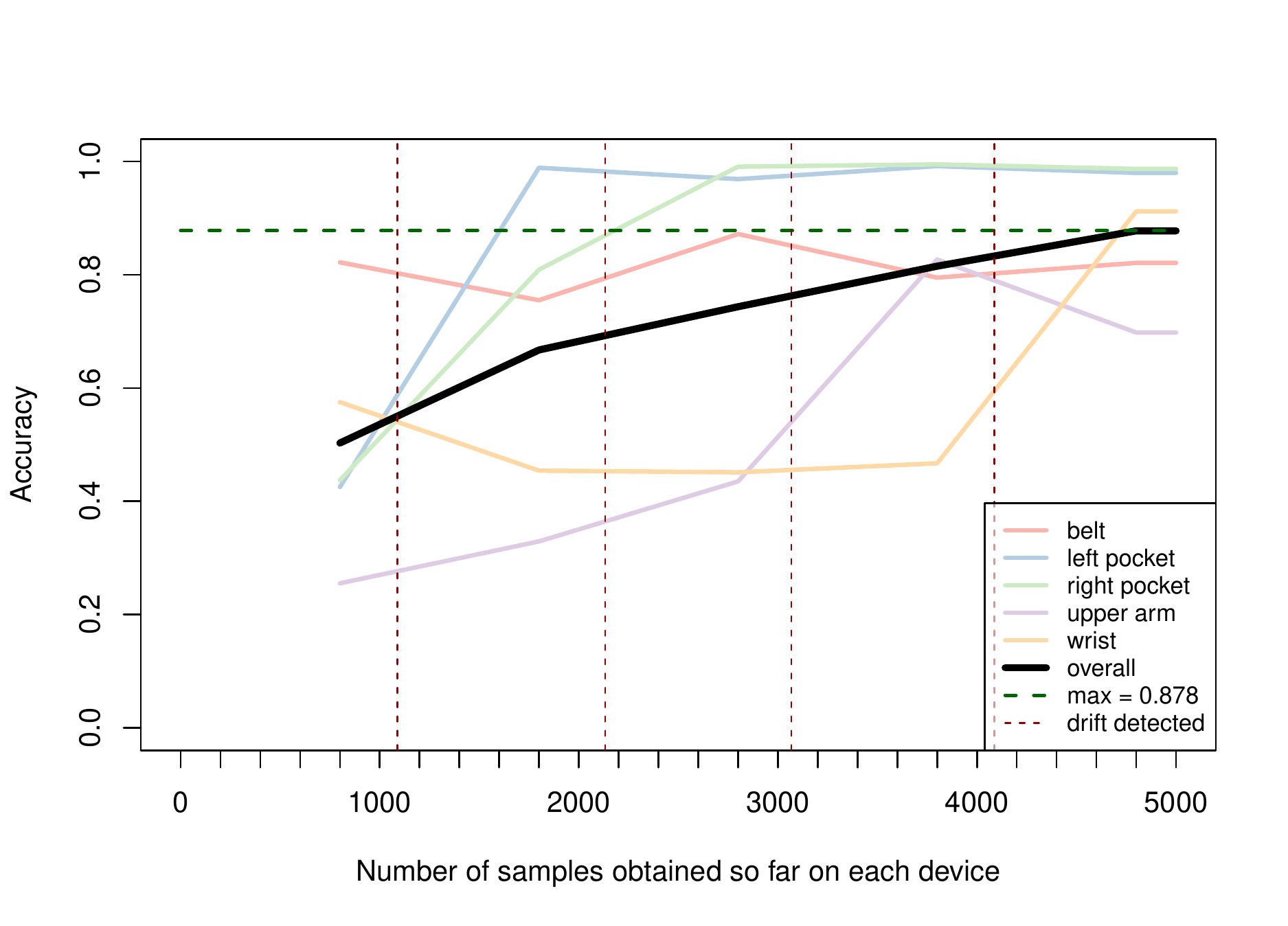}
	\caption{Results for \ourAcronym{} in a non-IID and nonstationary setting, training with all users except user 8, whose data is reserved for testing. The vertical dashed lines indicate when a drift is detected by our algorithm.}
	\label{fig:cda-fedavg-drifts1}
\end{figure}

\begin{figure}[htbp]
	\centering
	\includegraphics[width=0.9\textwidth,trim={0 10 15 50},clip]{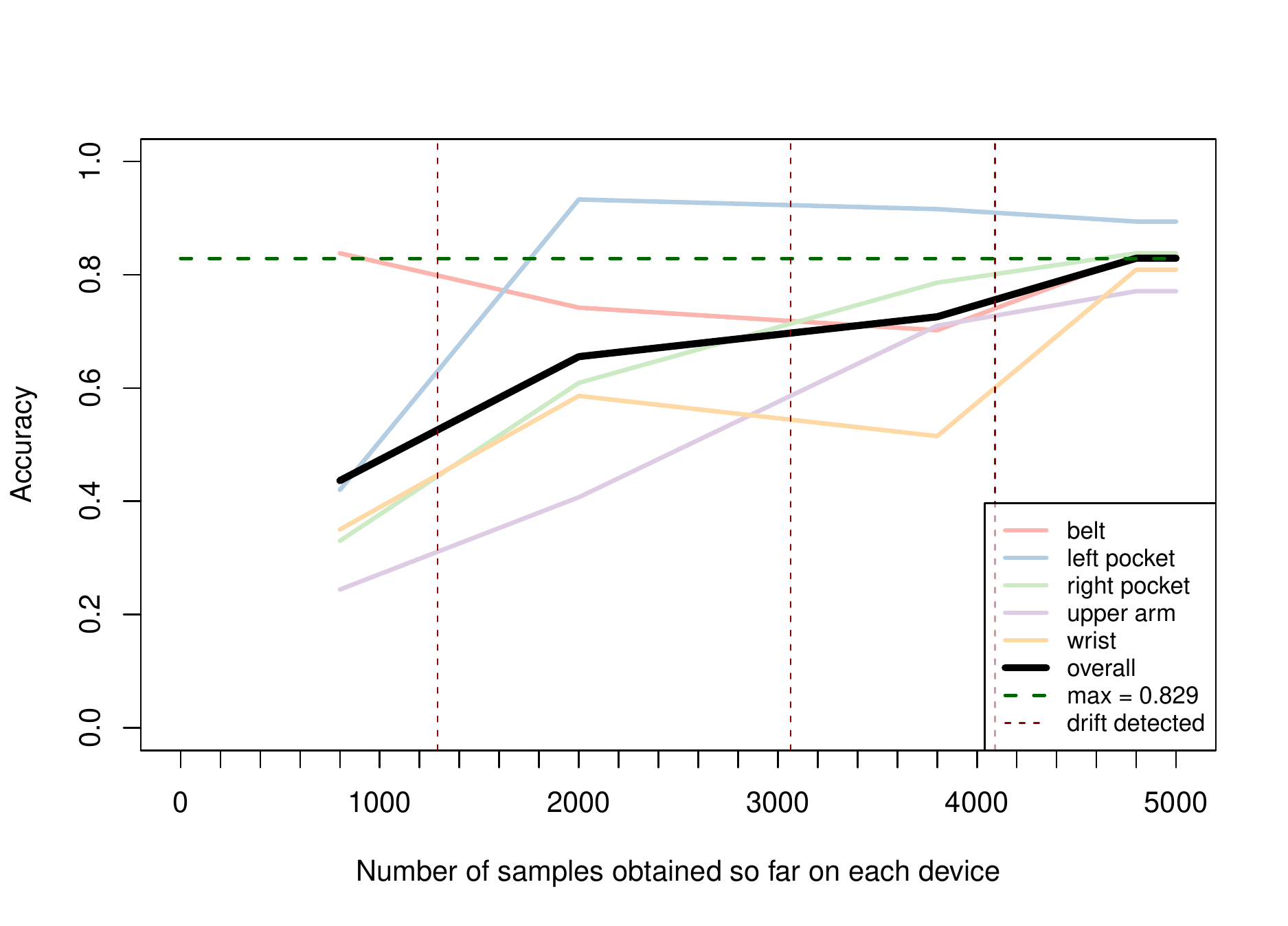}
	\caption{Results for \ourAcronym{} in a non-IID and nonstationary setting, training with all users except user 3, whose data is reserved for testing. The vertical dashed lines indicate when a drift is detected by our algorithm.} 
	\label{fig:cda-fedavg-drifts2}
\end{figure}

Finally, we repeated the last experiment but using our method, \ourAcronym{}, instead of regular FedAvg. This time, the training process is aware of the existence of the concept drifts. Unlike in the previous experiments, we cannot represent in a single plot the average evolution of the accuracy for the 10 executions, since in this case in each execution the drifts will be detected at slightly different times. Therefore, Figures~\ref{fig:cda-fedavg-drifts1}~and~\ref{fig:cda-fedavg-drifts2} show two particular executions to serve as examples. Table~\ref{tab:cda-fedavg-drifts} shows the results of all the 10 executions at the end of the process.

\begin{table}[htbp]
\centering
\caption{Final results for all the executions of \ourAcronym{} in a non-IID and nonstationary setting, after all clients have seen 5000 instances.}
\label{tab:cda-fedavg-drifts}
\resizebox{\textwidth}{!}{%
\begin{tabular}{l|cccccc}
\hline\noalign{\smallskip}
Test set & Belt  & Left pocket & Right pocket & Upper arm & Wrist & Overall \\
\noalign{\smallskip}\hline\noalign{\smallskip}
User 1   & 0.832 & 0.950       & 0.777        & 0.771     & 0.864 & 0.839   \\
User 2   & 0.786 & 0.955       & 0.669        & 0.691     & 0.797 & 0.780   \\
User 3   & 0.834 & 0.894       & 0.838        & 0.771     & 0.809 & 0.829   \\
User 4   & 0.936 & 0.983       & 0.971        & 0.613     & 0.908 & 0.884   \\
User 5   & 0.972 & 0.990       & 0.992        & 0.855     & 0.676 & 0.897   \\
User 6   & 0.776 & 0.941       & 0.722        & 0.712     & 0.908 & 0.812   \\
User 7   & 0.585 & 0.668       & 0.969        & 0.864     & 0.909 & 0.799   \\
User 8   & 0.821 & 0.980       & 0.987        & 0.698     & 0.912 & 0.878   \\
User 9   & 0.294 & 0.964       & 0.982        & 0.591     & 0.884 & 0.743   \\
User 10  & 0.748 & 0.749       & 0.561        & 0.684     & 0.883 & 0.725   \\
Average  & 0.759 & 0.908       & 0.847        & 0.726     & 0.855 & 0.819   \\
\noalign{\smallskip}\hline
\end{tabular}%
}
\end{table}

In this case, training does not start at the very beginning. Instead, the first round is not performed until every client has a representative amount of data from the first concept (here, belt position) in its long-term memory. We fix the number of training rounds per concept to 5. Thus, as there are 5 different changes, the model is trained for a total of 25 rounds. This has been intentionally designed to be on a par with the previous setting. Drifts occur at the same time points as in the previous case~(iterations 1000, 2000, 3000, and 4000). Nevertheless, now the vertical dashed lines indicate where a drift is actually detected by at least one of the clients applying our detection method. We can notice that all drifts are detected shortly after they theoretically occur. In some of the executions, such as the one shown in Figure~\ref{fig:cda-fedavg-drifts2}, the second drift is not detected at all. This makes sense, since this change is between the concepts of left pocket and right pocket, which are very similar. In case it was necessary to be more sensitive to change, it would be sufficient to vary the $\lambda$ parameter of Algorithm~\ref{alg:drift-detection}.

As shown in Table~\ref{tab:cda-fedavg-drifts}, using our method, the overall accuracy at the end of the learning process on the test set is around 82\%. This is much closer to the result obtained with the baseline model (Table~\ref{tab:fedavg-random}). Therefore, we can confirm that \ourAcronym{} is able to adapt to changes in nonstationary situations, while retaining previously learned concepts.

\section{Conclusions}
\label{sec:conclusions}
In this paper, we have tackled the problem of federated and continual learning under concept drift. We have started by discussing the issues that need to be addressed to achieve real multi-device learning. We have also reviewed the state of the art of continual and federated frameworks. 
We have shown the shortcomings of regular federated algorithms, such as FedAvg, when data is nonstationary over time, which is a common real-world situation. Therefore, we have developed a new method, \ourMethod{} (\ourAcronym{}).

Our proposal is an extension of the original FedAvg, but capable of detecting concept drifts and adapting to them. For drift detection, we introduce a distribution-based algorithm, which uses a confidence metric to quantify the dissimilarity between the distribution of historical and new data. We define a short-term and a long-term memory for each client. When a drift happens, we adapt the federated model applying rehearsal using the data in the long-term memory. In this way, we avoid catastrophic forgetting. Furthermore, we answer two fundamental questions: \emph{what} to learn and \emph{when} to learn it. This allows us to save storage, communication and computational resources. We have evaluated \ourAcronym{} in a real multiclass classification problem, human activity recognition, and we have shown that our method outperforms regular FedAvg in this kind of scenario.

Regarding our future work, we would like to continue to pursue this line of research. We want to further extend our framework for federated and continual learning, focusing on adaptability. Therefore, we will work in parallel in two dimensions: the temporal and the spatial. The temporal dimension is the one we have explored the most so far. However, we can still make progress on issues such as tackling not only the virtual but also the real concept drift. When we talk about the spatial dimension, we refer to the heterogeneity among clients and the adaptation to the local particularities of each one. There are already some proposals in this context, but they are still limited by certain assumptions that make them unsuitable for real-world applications. For instance, it is not considered that different clients may have different behaviors and therefore could label the same pattern in a distinct way. Nevertheless, we believe that the real challenge lies in addressing these two axes, temporal and spatial, at the same time. This is something that has not yet been done and opens up a much richer and more complex range of possibilities.  Finally, we also want to gradually expand our experimental analysis, applying our algorithms to other applications in different fields, not only for smartphones. We are particularly interested in service robotics.

\begin{acknowledgements}
This research has received financial support from AEI/FEDER (EU) grant number TIN2017-90135-R, as well as the \textit{Conseller\'ia de Cultura, Educaci\'on e Ordenaci\'on Universitaria} of Galicia (accreditation 2016--2019, ED431G/01 and ED431G/08, reference competitive group ED431C2018/29, and grant ED431F2018/02), and the European Regional Development Fund (ERDF). It has also been supported by the \textit{Ministerio de Universidades} of Spain in the FPU 2017 program (FPU17/04154).
\end{acknowledgements}

\section*{Conflict of interest}
The authors declare that they have no conflict of interest.

%
%
\bibliographystyle{spmpsci_unsrt}
\bibliography{MTAP_drift}

\begin{thebibliography}{10}
\providecommand{\url}[1]{{#1}}
\providecommand{\urlprefix}{URL }
\expandafter\ifx\csname urlstyle\endcsname\relax
  \providecommand{\doi}[1]{DOI~\discretionary{}{}{}#1}\else
  \providecommand{\doi}{DOI~\discretionary{}{}{}\begingroup
  \urlstyle{rm}\Url}\fi

\bibitem{konevcny2015federated}
Kone{\v{c}}n{\`y}, J., McMahan, B., Ramage, D.: Federated optimization:
  Distributed optimization beyond the datacenter.
\newblock arXiv preprint arXiv:1511.03575  (2015)

\bibitem{mcmahan2016federated}
McMahan, H.B., Moore, E., Ramage, D., Aguera-Arcas, B.: Federated learning of
  deep networks using model averaging.
\newblock arXiv preprint arXiv:1602.05629v1  (2016)

\bibitem{li2019federated}
Li, Q., Wen, Z., He, B.: Federated learning systems: Vision, hype and reality
  for data privacy and protection.
\newblock arXiv preprint arXiv:1907.09693  (2019)

\bibitem{custers2019eu}
Custers, B., Sears, A.M., Dechesne, F., Georgieva, I., Tani, T., van~der Hof,
  S.: EU Personal Data Protection in Policy and Practice.
\newblock Springer (2019)

\bibitem{gaff2014privacy}
Gaff, B.M., Sussman, H.E., Geetter, J.: Privacy and big data.
\newblock Computer \textbf{47}(6), 7--9 (2014)

\bibitem{lu2018}
Lu, J., Liu, A., Dong, F., Gu, F., Gama, J., Zhang, G.: Learning under concept
  drift: A review.
\newblock IEEE Transactions on Knowledge and Data Engineering  (2018)

\bibitem{lesort2020continual}
Lesort, T., Lomonaco, V., Stoian, A., Maltoni, D., Filliat, D.,
  D{\'\i}az-Rodr{\'\i}guez, N.: Continual learning for robotics: Definition,
  framework, learning strategies, opportunities and challenges.
\newblock Information Fusion \textbf{58}, 52--68 (2020)

\bibitem{parisi2019}
Parisi, G.I., Kemker, R., Part, J.L., Kanan, C., Wermter, S.: Continual
  lifelong learning with neural networks: A review.
\newblock Neural Networks  (2019)

\bibitem{grossberg1988}
Grossberg, S.: Nonlinear neural networks: Principles, mechanisms, and
  architectures.
\newblock Neural networks \textbf{1}(1), 17--61 (1988)

\bibitem{casado2020concept}
Casado, F.E., Lema, D., Iglesias, R., Regueiro, C.V., Barro, S.: Concept drift
  detection and adaptation for robotics and mobile devices in federated and
  continual settings.
\newblock In: Workshop of Physical Agents, pp. 79--93. Springer (2020)

\bibitem{french1999catastrophic}
French, R.M.: Catastrophic forgetting in connectionist networks.
\newblock Trends in cognitive sciences \textbf{3}(4), 128--135 (1999)

\bibitem{kairouz2019advances}
Kairouz, P., McMahan, H.B., Avent, B., Bellet, A., Bennis, M., Bhagoji, A.N.,
  Bonawitz, K., Charles, Z., Cormode, G., Cummings, R., et~al.: Advances and
  open problems in federated learning.
\newblock arXiv preprint arXiv:1912.04977  (2019)

\bibitem{li2020federated}
Li, T., Sahu, A.K., Talwalkar, A., Smith, V.: Federated learning: Challenges,
  methods, and future directions.
\newblock IEEE Signal Processing Magazine \textbf{37}(3), 50--60 (2020)

\bibitem{aledhari2020federated}
Aledhari, M., Razzak, R., Parizi, R.M., Saeed, F.: Federated learning: A survey
  on enabling technologies, protocols, and applications.
\newblock IEEE Access \textbf{8}, 140,699--140,725 (2020)

\bibitem{zhao2018federated}
Zhao, Y., Li, M., Lai, L., Suda, N., Civin, D., Chandra, V.: Federated learning
  with non-iid data.
\newblock arXiv preprint arXiv:1806.00582  (2018)

\bibitem{caldas2018leaf}
Caldas, S., Wu, P., Li, T., Kone{\v{c}}n{\`y}, J., McMahan, H.B., Smith, V.,
  Talwalkar, A.: Leaf: A benchmark for federated settings.
\newblock arXiv preprint arXiv:1812.01097  (2018)

\bibitem{li2018federated}
Li, T., Sahu, A.K., Zaheer, M., Sanjabi, M., Talwalkar, A., Smith, V.:
  Federated optimization in heterogeneous networks.
\newblock arXiv preprint arXiv:1812.06127  (2018)

\bibitem{deng2020adaptive}
Deng, Y., Kamani, M.M., Mahdavi, M.: Adaptive personalized federated learning.
\newblock arXiv preprint arXiv:2003.13461  (2020)

\bibitem{yoon2020federated}
Yoon, J., Jeong, W., Lee, G., Yang, E., Hwang, S.J.: Federated continual
  learning with weighted inter-client transfer.
\newblock arXiv preprint arXiv:2003.03196v4  (2020)

\bibitem{casado2020collaborative}
Casado, F.E., Lema, D., Iglesias, R., Regueiro, C.V., Barro, S.: Collaborative
  and continual learning for classification tasks in a society of devices.
\newblock arXiv preprint arXiv:2006.07129v2  (2020)

\bibitem{webb2016characterizing}
Webb, G.I., Hyde, R., Cao, H., Nguyen, H.L., Petitjean, F.: Characterizing
  concept drift.
\newblock Data Mining and Knowledge Discovery \textbf{30}(4), 964--994 (2016)

\bibitem{gepperth2016incremental}
Gepperth, A., Hammer, B.: Incremental learning algorithms and applications.
\newblock In: Proceedings of the European Symposium on Artificial Neural
  Networks, Computational Intelligence and Machine Learning (ESANN), pp.
  357--368. i6doc (2016)

\bibitem{baron1999convergence}
Baron, M.: Convergence rates of change-point estimators and tail probabilities
  of the first-passage-time process.
\newblock Canadian Journal of Statistics \textbf{27}(1), 183--197 (1999)

\bibitem{haque2016}
Haque, A., Khan, L., Baron, M.: Sand: Semi-supervised adaptive novel class
  detection and classification over data stream.
\newblock In: Thirtieth AAAI Conference on Artificial Intelligence, pp.
  1652--1658 (2016)

\bibitem{bowman2014estimation}
Bowman, K., Shenton, L.: Estimation: Method of moments.
\newblock Wiley StatsRef: Statistics Reference Online  (2014)

\bibitem{van2019three}
van~de Ven, G.M., Tolias, A.S.: Three scenarios for continual learning.
\newblock arXiv preprint arXiv:1904.07734  (2019)

\bibitem{armijo1966minimization}
Armijo, L.: Minimization of functions having lipschitz continuous first partial
  derivatives.
\newblock Pacific Journal of mathematics \textbf{16}(1), 1--3 (1966)

\bibitem{konevcny2016federated}
Kone{\v{c}}n{\`y}, J., McMahan, H.B., Yu, F.X., Richt{\'a}rik, P., Suresh,
  A.T., Bacon, D.: Federated learning: Strategies for improving communication
  efficiency.
\newblock arXiv preprint arXiv:1610.05492  (2016)

\bibitem{hard2018federated}
Hard, A., Rao, K., Mathews, R., Ramaswamy, S., Beaufays, F., Augenstein, S.,
  Eichner, H., Kiddon, C., Ramage, D.: Federated learning for mobile keyboard
  prediction.
\newblock arXiv preprint arXiv:1811.03604  (2018)

\bibitem{shoaib2014fusion}
Shoaib, M., Bosch, S., Incel, O.D., Scholten, H., Havinga, P.J.: Fusion of
  smartphone motion sensors for physical activity recognition.
\newblock Sensors \textbf{14}(6), 10,146--10,176 (2014)

\bibitem{casado2020walking}
Casado, F.E., Rodr{\'\i}guez, G., Iglesias, R., Regueiro, C.V., Barro, S.,
  Canedo-Rodr{\'i}guez, A.: Walking recognition in mobile devices.
\newblock Sensors \textbf{20}(4) (2020)

\bibitem{tong2021cnn}
Tong, L.N., He, J.J., Peng, L.: {CNN-based PD hand tremor detection using
  inertial sensor}.
\newblock IEEE Sensors Letters  (2021)

\end{thebibliography}
\end{document}